**Invitation to tender ESA/AO/1-8373/15/I-NB – "VAE: Next Generation EO-based Information Services"**

# Automatic Spatial Context-Sensitive Cloud/Cloud-Shadow Detection in Multi-Source Multi-Spectral Earth Observation Images –

# AutoCloud+

**Author: Andrea Baraldi, Senior scientist: ……………………**

**Dept. of Agricultural and Food Sciences, University of Naples Federico II, Portici (NA), Italy (e-mail: andrea6311@gmail.com)**
**Paris Lodron University of Salzburg (PLUS), Austria / Department of Geoinformatics (Z_GIS)**

**Date: Nov. 3, 2015.**

**Prof. Dr. Josef Strobl, Head of Z_GIS-PLUS: …………………**

***ESA TENDERING STANDARDS FOR***
***"EXPRESS PROCUREMENT" ("EXPRO" & "EXPRO+")***

**AUTOMATIC SPATIAL CONTEXT-SENSITIVE CLOUD/CLOUD-SHADOW DETECTION IN MULTI-SOURCE MULTI-SPECTRAL EARTH OBSERVATION IMAGES:**
**AUTOCLOUD+**

# Contents

# 1 TECHNICAL PART

The proposed Earth observation (EO)-based value adding system (EO-VAS), hereafter identified as AutoCloud+, consists of an innovative EO image understanding system (EO-IUS) design and implementation capable of automatic spatial context-sensitive cloud/cloud-shadow detection in multi-source multi-spectral (MS) EO imagery, whether or not radiometrically calibrated, acquired by multiple platforms, either spaceborne or airborne, including unmanned aerial vehicles (UAVs). It is worth mentioning that the same EO-IUS architecture is suitable for a large variety of EO-based value-adding products and services, including: (i) low-level image enhancement applications, such as automatic MS image topographic correction, co-registration, mosaicking and compositing, (ii) high-level MS image land cover (LC) and LC change (LCC) classification and (iii) content-based image storage/retrieval in massive multi-source EO image databases ("big data" mining).

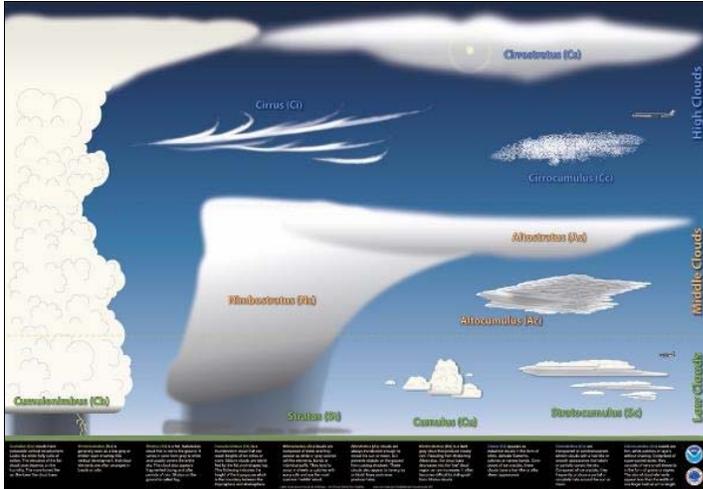

Figure 1. Cloud classification according to the U.S. National Weather Service (www.srh.noaa.gov/jetstream/clouds/corefour.htm).

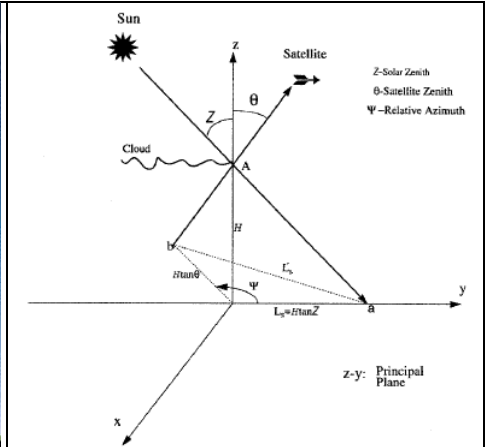

Figure 2. Sun-cloud satellite geometry (leading cloud edge, point A) for arbitrary viewing and illumination conditions. H and LS are the actual cloud height and cloud shadow length. LS' is the apparent cloud shadow length.

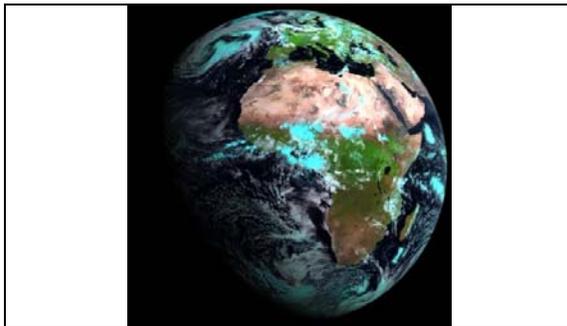

Figure 3(a). Quick-look RGB image of a Meteosat 2nd Generation (MSG) SEVIRI image acquired on 2012-05-30, 08:45, radiometrically calibrated into top-of-atmosphere reflectance (TOARF) values and depicted in false colors (R: band 3, G: band 2, B: band 1), spatial resolution: 3 km. A default ENVI 2% linear histogram stretching was applied for visualization purposes.

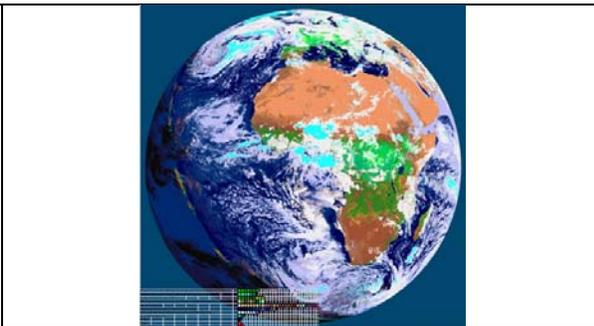

Figure 3(b). 4-band AVHRR-like Satellite Image Automatic Mapper (AV-SIAM)'s preliminary color map, featuring 83 color names depicted in pseudocolors, automatically generated from the MSG image shown in Figure 3(a). Map legend: shown in the lower left corner. For greater details about the SIAM's map legends, refer to text.

## 1.1 EO-VAS OBJECTIVES, TECHNICAL REQUIREMENTS AND PROPOSED APPROACH

Since the proposed EO-VAS for cloud/cloud-shadow detection in multi-source MS images has not been published yet, the present project proposal as well as any further documentation regarding the proposed activity shall be regarded as "Proprietary Sensitive Information", subject to Articles 6.1.2 and 6.1.3 of the ESA Contract No. 4000xxxxxx/15/I-NB, ARTICLE 6 -Information to be provided by the Contractor – Protection of information, whose quotes are the following.

- "6.1.2 For the purpose of this Contract Proprietary Sensitive Information… The Contractor shall not mark any (electronic) documentation as Proprietary Sensitive Information, unless agreed in advance with the Agency. Any request from the Contractor shall be submitted in writing accompanied by an appropriate justification."

- "6.1.3 Neither Party shall disclose any documentation obtained from the other Party, and which both Parties recognise as being Proprietary Sensitive Information without the other Party's previous written authorisation."

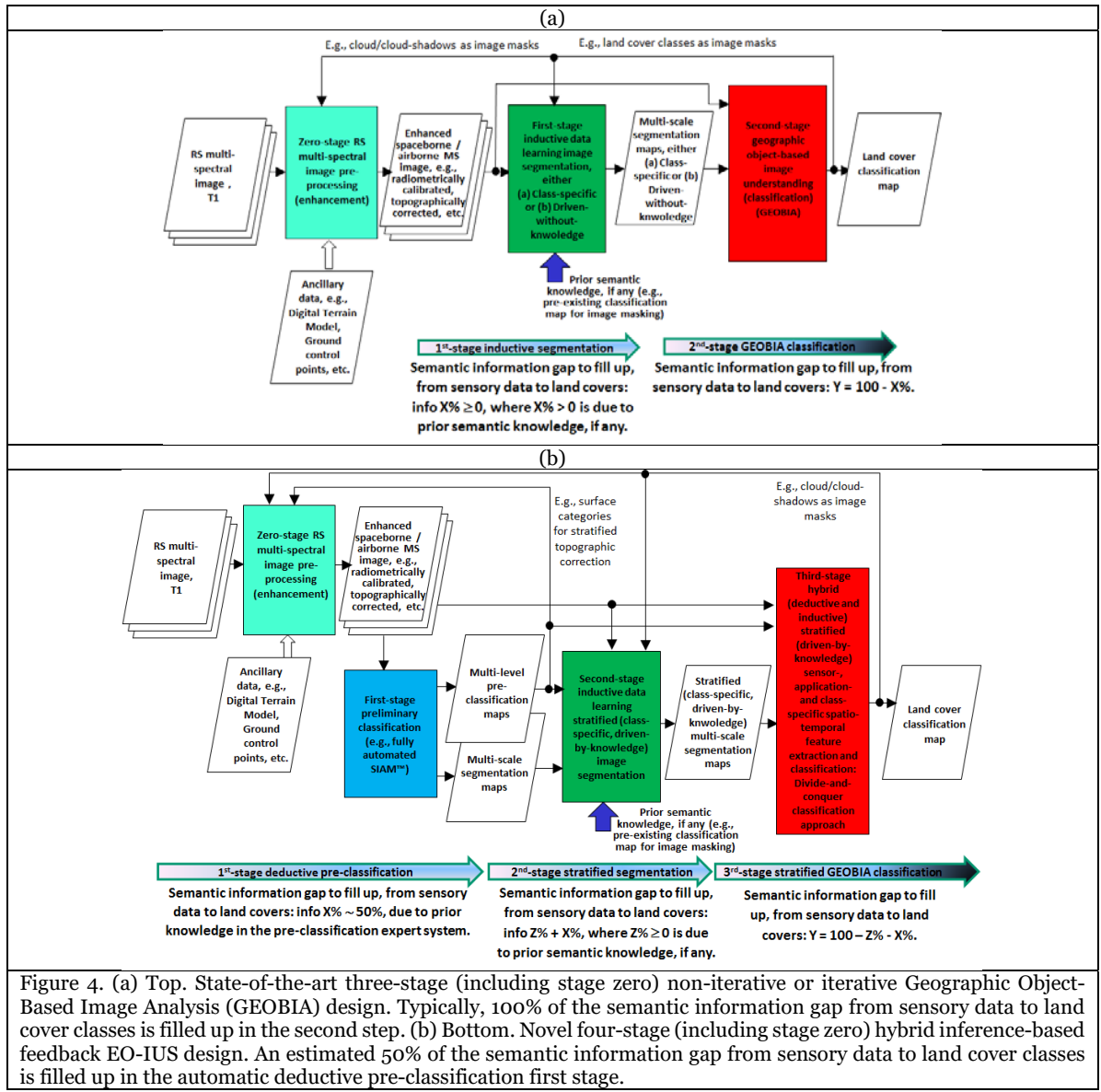

Figure 4. (a) Top. State-of-the-art three-stage (including stage zero) non-iterative or iterative Geographic Object-Based Image Analysis (GEOBIA) design. Typically, 100% of the semantic information gap from sensory data to land cover classes is filled up in the second step. (b) Bottom. Novel four-stage (including stage zero) hybrid inference-based feedback EO-IUS design. An estimated 50% of the semantic information gap from sensory data to land cover classes is filled up in the automatic deductive pre-classification first stage.

### 1.1.1 EO-VAS aims and degrees of innovation

In the remote sensing (RS) community, a well-known prerequisite for clear-sky RS image compositing [1]-[5], suitable for further retrieval of land surface variables, either quantitative [6], such as biophysical variables, e.g., the leaf area index (LAI), or categorical (nominal) variables [6], such as LC/LCC classes, is accurate masking of clouds and cloud shadows, see Figure 1 to Figure 3. Intuitively, cloud contamination is a relevant problem in LCC analysis, because unflagged clouds may be mapped as false LCC occurrences.

In compliance with the Quality Assurance Framework for Earth Observation (QA4EO) guidelines, developed by the intergovernmental Group on EOs (GEO) [7], the ambitious goal of the present software project is to undertake the first research and technological development (RTD) of an EO-IUS software pipeline capable of cloud and cloud shadow detection in one input EO multi-source MS image subject to the following RTD project requirements specification. The proposed EO-IUS must be: (I) automatic, i.e., it requires no user's interaction. (II) In operating mode, i.e., ready-for-use, by scoring high in a set of metrological/statistically-based quantitative quality indicators (Q2Is) of operativeness (Q2IOs), to be RS community-agreed upon. Proposed Q2IOs to be jointly maximized encompass [8], [9]: (i) degree of automation, (ii) accuracy, (iii) efficiency, (iv) robustness to changes in input parameters, (v) robustness to changes in input data, (vi) scalability/transferability, (vii) timeliness, from data acquisition to data-derived product generation, and (viii) economy (vice versa, costs in manpower and computer power must be kept low). Noteworthy, this set of Q2IOs has never been adopted in the RS literature on a regular basis. (III) Sensor-independent, which means: (a) multi-scale, from regional to global spatial extents, (b) multi-resolution, from coarse (≈ 1 km) to very high (< 1 m), and (c) multi-platform, either spaceborne or airborne, including unmanned aerial vehicles (UAVs) [10]. (IV) Input with an MS image that is either: (a) radiometrically calibrated into top-of-atmosphere (TOA) reflectance (TOARF) or surface reflectance (SURF) values [11], or (b) uncalibrated. Although highly recommended in compliance with the QA4EO guidelines [7], radiometric calibration of digital numbers (DNs) is not considered mandatory by the present RTD project [11], to cope with consumer-level color cameras typically mounted on board light-weight UAVs or terrestrial photocameras [10].

Conceived to outperform existing state-of-the-art cloud/cloud-shadow detectors (see Figure 4(a)), which are typically semi-automatic, site-specific and sensor-dependent (see Table 3), the aforementioned RTD software project's requirements, (I) to (IV), are ambitious, but realistic. They rely upon several EO-IUS software units already implemented, tested and validated by third-parties in recent years [8], [9], [12]-[14]. These software units can be combined according to an innovative EO-IUS architecture, see Figure 4(b), featuring hybrid inference mechanisms (combined deductive/top-down/prior knowledge-based inference with inductive/bottom-up/learning-from-data inference) and provided with feedback loops. According to Marr [15], the linchpin of success of any information processing system is addressing the level of understanding of computational theory (system design), rather than algorithms or implementation. Noteworthy, in the EO-IUS design proposed in Figure 4(b), first-stage prior knowledge-based inference (analogous to genotype) predates and conditions second-stage inductive data learning (equivalent to phenotype), because the latter is inherently ill-posed and requires *a priori* knowledge in addition to data to become better posed for numerical solution [21]. Feedback loops allow to back-project high-level categorical variables onto input quantitative variables, to accomplish either data enhancement, such as automatic EO image topographic correction [35] (see Figure 5), or stratified (masked, conditioned) biophysical variable estimation, e.g., LAI estimation [54]. The hybrid feedback EO-IUS architecture shown in Figure 4(b) is alternative to that adopted by the mainstream RS community, whose EO-IUSs adopt a feedforward inductive data learning strategy.

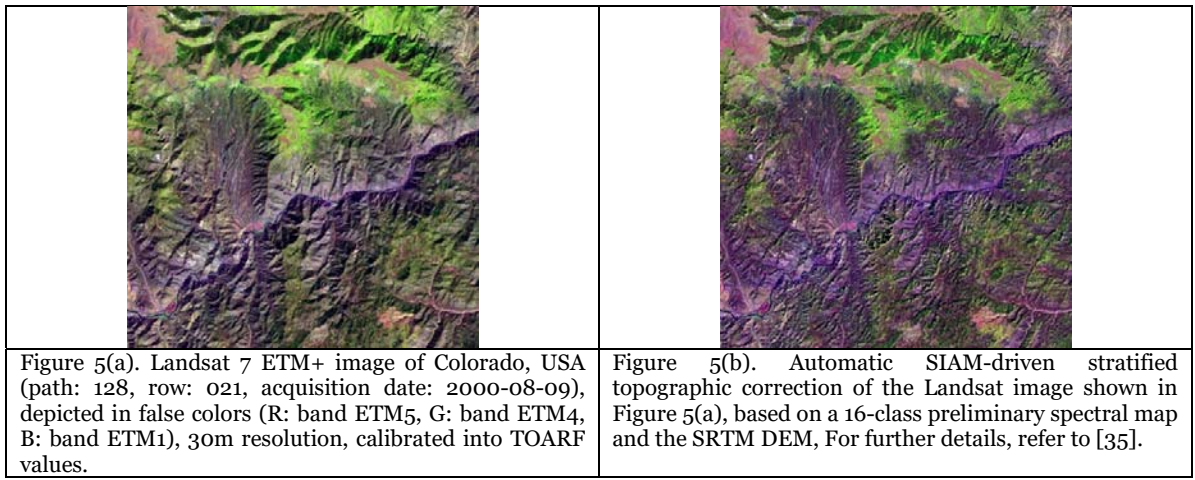

| Figure 5(a). Landsat 7 ETM+ image of Colorado, USA (path: 128, row: 021, acquisition date: 2000-08-09), depicted in false colors (R: band ETM5, G: band ETM4, B: band ETM1), 30m resolution, calibrated into TOARF values. | Figure 5(b). Automatic SIAM-driven stratified topographic correction of the Landsat image shown in Figure 5(a), based on a 16-class preliminary spectral map and the SRTM DEM, For further details, refer to [35]. |

In addition to being considered realistic, the present RTD software project can be assessed as potentially relevant for the whole RS community. If successful, it would provide the first proof-of-concept that the proposed novel EO-IUS design and implementation strategies are capable of transforming multi-source EO image "big data" into operational, comprehensive and timely information products, e.g., cloud/cloud-shadow masks, in compliance with the QA4EO recommendations and with several ongoing RS international programs [16], [17]. This first proof-of-concept would open a wide spectrum of future research, educational and market opportunities, including the following.

(A) Accomplish automatic estimation of either continuous variables, such as topographically corrected TOARF values [35] (see Figure 5) or biophysical variables, e.g., LAI [54], or categorical variables, such as LC/LCC classes, from existing massive multi-sensor EO image repositories [46], to better understand the systemic and interrelated nature of global LC/LCC dynamics.

(B) Integrate near real-time internet-based satellite mapping services on demand with virtual Earth geo-browsers, such as the popular Google Earth, see Figure 6.

(C) Augment the scientific and commercial impact of European space infrastructures, including the Sentinel-2 MSI, Sentinel-3 OLCI and SLSTR imaging sensors, the Meteosat satellite 1st, 2nd and 3rd generation sensor series, etc., because the adopted expert system for MS image pre-classification and segmentation, the Satellite Image Automatic Mapper (SIAM) [9], [12]-[14], is capable of mapping any radiometrically calibrated MS image acquired by past, existing or future MS imaging sensors [18], e.g., Envisat AATSR, Meris, SPOT-1/2/3/4/5, SPOT-6/7, Pleiades-1A/B, IRS-1C/D, IRS-P6, AVHRR, Modis, Landsat-4/5/7/8, World View-2/3, Ikonos, QuickBird, GeoEye-1, RapidEye, Skybox, Planetlabs, Alos-1/2, etc. Noteworthy, the SIAM pre-classification maps available to date, see Figure 3 and Figure 7, are more informative than the future Sentinel-2A Level 2 products, consisting of a cloud mask and a land/water mask, to be generated on a non-systematic basis, and of the Landsat-8 quality bands, consisting of cloud masks, already available on a regular basis.

(D) Integrate the visual analysis of uncalibrated RGB images acquired by consumer-level terrestrial and aerial color cameras, such as those mounted onboard light-weight UAVs, onto the same EO-IUS pipeline adopted for the interpretation of MS images radiometrically calibrated into TOARF or SURF values in compliance with the QA4EO guidelines, see Figure 8.

(E) The automatic extraction of content maps from EO imagery allows each EO image stored in a massive EO image database to be provided with one or more content maps. The solution of the problem of automatic image-derived content map extraction guarantees the solution of its dual problem, specifically, content-based image storage/retrieval [46], which is an open problem to date. The latter would become a seamless navigation through content maps where image-objects (segments, patches, polygons) provided with semantic labels can be tracked through time [51].

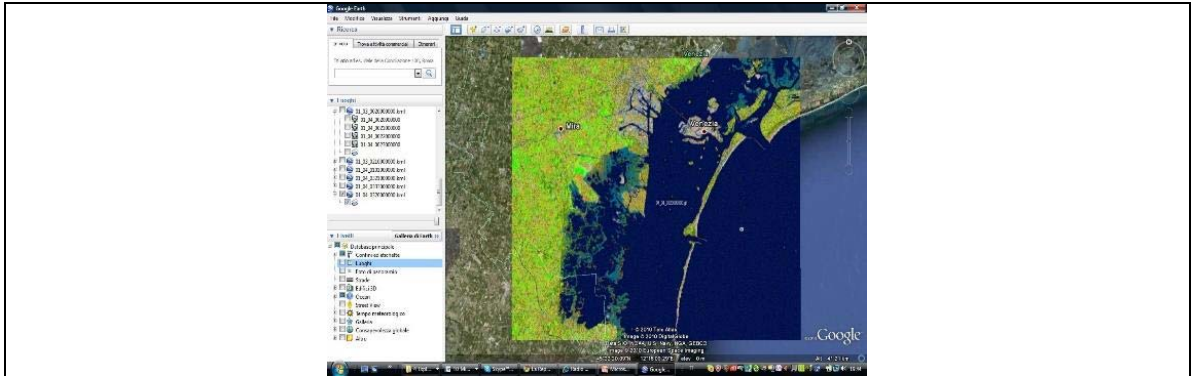

Figure 6. Preliminary classification map, depicted in pseudocolors, generated by the SIAM expert system from a Landsat 7 ETM+ image of the Venice lagoon, Italy, radiometrically calibrated into TOARF values, spatial resolution: 30 m. The SIAM map was transformed into the .kml data format and uploaded as a thematic layer in a commercial 3-D Earth viewer (e.g., Google Earth).

### 1.1.2 EO-VAS architecture and implementation

To comply with the EO-IUS requirements specified in Section 1.1.1 and overcome operational limitations of the existing cloud/cloud-shadow detectors listed in Table 3, to be further discussed in Section 1.1.4, an original implementation of the four-stage EO-IUS architecture, shown in Figure 4(b), is proposed in agreement with the software pipeline sketched in Figure 9.

**I. Spaceborne/airborne MS image pre-processing, identified as Stage 0 (zero) in Figure 4(b).**

In compliance with the QA4EO guidelines [7], the proposed RS image pre-processing Stage 0 must include the radiometric calibration of DNs into TOARF, SURF or surface albedo values, where TOARF $\supseteq$ SURF (i.e., SURF is a special case of TOARF in flat terrain and very clear sky conditions [28]). On theory, the RS community regards as common knowledge the prerequisite that for physically based, quantitative analysis of airborne and satellite sensor measurements in the optical domain their calibration to spectral radiance or reflectance values is mandatory [7]. Unfortunately, in the RS common practice, scientists, practitioners and institutions tend to overlook radiometric calibration as a necessary not sufficient pre-processing requirement capable of harmonizing large-scale multi-temporal multi-sensor EO datasets. Physical model-based and hybrid (combined physical and statistical) EO-IUSs do require as input sensory data provided with a physical unit of radiometric measure. On the contrary, statistical model-based EO-IUSs do not require as input numerical variables subject to radiometric calibration, e.g., refer to Table 3. Nonetheless, statistical systems too can benefit from the harmonization of input data accomplished via radiometric calibration. It is worth mentioning that the present RTD software project's background relies, free of cost, on an existing battery of computer programs for sensor-specific MS image radiometric calibration of DNs into TOARF values.

**II. Physical model-based per-pixel pre-attentive vision first stage for MS image multi-granule pre-classification and multi-scale segmentation, identified as Stage 1 in Figure 4(b).**

Although rarely acknowledged by the RS community, prior knowledge-based pre-classification for MS data space discretization has a long history. For example, it is part of the atmospheric correction implemented in the ATCOR commercial software product [25]. It is also part of the NASA and the Canadian Centre for Remote Sensing (CCRS) automatic processing chain for MODIS data composites [1], [29]. Finally, Shackelford and Davis adopted a statistical model-based (maximum likelihood) pre-classification first stage to stratify (partition) a second-stage battery of LC class-specific feature extractors and classification modules [30], [31]. Equivalent to color naming in a natural language [32], prior knowledge-based color space discretization is the deductive counterpart of popular unsupervised (unlabeled) data learning algorithms for vector quantization (VQ) [21], like the popular k-means VQ algorithm, not to be confused with unsupervised data clustering algorithms [33].

To be input with a MS image whether or not radiometrically calibrated, in compliance with Section 1.1.1, the proposed EO-IUS for cloud/cloud-shadow detection, sketched in Figure 4(b), adopts an original implementation of the pre-attentive vision Stage 1. It consists of two existing prior knowledge-based color data discretization algorithms in operating mode, the SIAM [8], [9], [12]-[14] (see Table 1) and the new RGB-Image Automatic Mapper (RGBIAM) [50] (see Table 2). When the input MS image is radiometrically calibrated, then both SIAM and RGBIAM can be run in parallel to combine color strata compatible with the presence of clouds and cloud shadows, according to a convergence-of-evidence approach. Otherwise, if the input image is uncalibrated then the sole RGBIAM can be employed. The two SIAM and RGBIAM expert systems are described below.

(i) The multi-source prior knowledge-based SIAM decision tree for MS data space discretization (color naming [32]) has been proposed, tested and validated by the RS community in recent years [8], [9], [12]-[14]. It is capable of generating, alternately automatically and in near real-time, multi-level pre-classification maps and multi-scale segmentation maps (computed from pre-classification maps via a well-posed two-pass connected-component multi-level image labeling algorithm [52]) of a spaceborne/airborne MS image radiometrically calibrated into TOARF or SURF values. At the level of abstraction of knowledge/information representation, the legend of a SIAM's pre-classification map generated from a single-date MS imagery consists of a discrete and finite vocabulary of *color names* [32], also called *spectral-based semi-concepts* or *spectral categories*, such as green-as-"*vegetation*", brown-as-"*bare soil or built-up*", blue-as-"*water or shadow*", etc. [8], [9], [12]-[14], see Table 1, Figure 3 and Figure 7. Each spectral-based semi-concept can match none, one or more target LC classes whose spectral properties overlap, irrespective of the other dominant spatio-temporal properties of these LC classes (e.g., "deciduous forest") in the 4-D real world-through-time. In other words, discrete color types, just like endmember fractions, "cannot always be inverted to unique LC class names" [27] (p. 147). At the level of understanding of system design, the SIAM software product is implemented as an integrated system of four subsystems, including one "master" 7-band Landsat-like (visible blue (B), visible green (G), visible red (R), near infra-red (NIR), medium infra-red 1 (MIR1), medium infra-red 2 (MIR2) and thermal infra-red (TIR)) subsystem plus three "slave" (downscale) subsystems, namely, a 4-band SPOT-like (G, R, NIR and MIR1), a 4-band AVHRR-like (R, NIR, MIR1 and TIR) and a 4-band VHR-like (B, G, R and NIR), whose spectral resolutions overlap with Landsat's, but are inferior to Landsat's. The expression "Landsat-like MS image" adopted in this paper means: "an MS image whose spectral resolution mimics the spectral domain of the 7 bands of the Landsat family of imaging sensors", i.e., a spectral resolution where bands B, G, R, NIR, MIR1, MIR2 and TIR overlap (which does not mean coincide) with Landsat's.

Table 1. Example of a preliminary classification map's legend, adopted by the 7-band Landsat-like SIAM subversion (L-SIAM) at fine semantic granularity, consisting of 96 spectral categories (color names). Pseudocolors of the spectral categories are grouped on the basis of their spectral end member (e.g., brown-as-"bare soil or built-up") or parent spectral category (e.g., "high" leaf area index (LAI) vegetation types). The pseudocolor of a spectral category is chosen so as to mimic natural colors of pixels belonging to that spectral category.

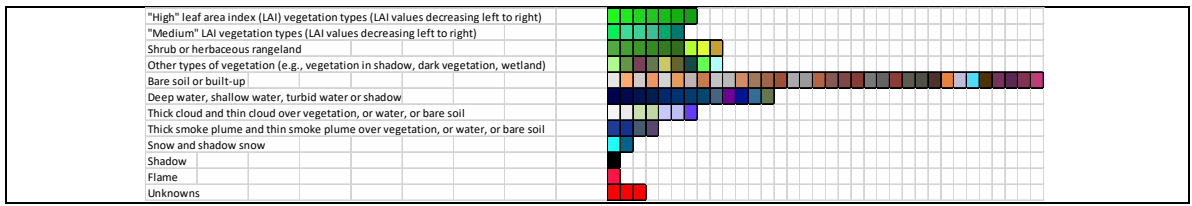

Table 2. Example of a preliminary classification map's legend adopted by the 3-band RGB Image Automatic Mapper (RGBIAM), suitable for preliminary classification of non-calibrated color images, consisting of 26 spectral categories (color names). Pseudocolors of the spectral categories are grouped on the basis of their spectral end member (e.g., brown-as-"bare soil or built-up"). The pseudocolor of a spectral category is chosen so as to mimic natural colors of pixels belonging to that spectral category.

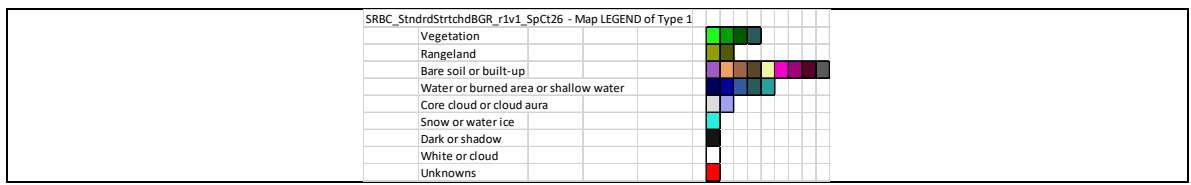

(ii) A recent extension of the SIAM software product, called RGBIAM [50], was developed for preliminary classification of an RGB (true color or false color) data space not subject to radiometric calibration, but submitted to an original inductive (data-driven) histogram stretching capable of accomplishing the so-called *color constancy effect* in human vision, i.e., color normalization under a "canonical" illumination source [34], refer to Table 2. Typically, an image band-specific histogram features up to three main modes: high (scene foreground, if any), central and low (scene background, if any). If the background and foreground histogram modes exist and are detected, then they are compressed at the opposite ends of the color domain, e.g., equal to 0 and 255 in byte data-coding, whereas the central mode is stretched linearly between these two end values. Hence, the per-band image contrast is enhanced and the adaptive color constancy algorithm accomplishes inter-image harmonization, equivalent to radiometric calibration. Finally, an RGBIAM expert system partitions an RGB data cube, subject to color constancy, onto a mutually exclusive and totally exhaustive finite and discrete set of fixed (non-adaptive to input data) color names (quantization levels), corresponding to 3-D polyhedral. An example of an RGBIAM color mapping applied to a non-calibrated Landsat-8 image is shown in Figure 8.

III. **Feedback loops, from the pre-attentional Stage 1 and the attentional Stage 3 into the pre-processing Stage 0, refer to Figure 4(b).**

This is where a relevant degree of novelty of the innovative four-stage EO-IUS architecture is located, see Figure 4(b). A first feedback loop feeds the pre-attentive vision Stage 1's categorical output back to the pre-processing Stage 0's input for stratified (driven-by-knowledge, symbolic mask-conditioned) automatic RS image enhancement. Existing examples are stratified atmospheric correction [25], stratified topographic correction [35] (see Figure 5), stratified image-co-registration and stratified image mosaic enhancement [14]. A second feedback loop feeds the attentive vision Stage 3's categorical output back to the pre-processing Stage 0's input, e.g., for cloud/cloud-shadow masking, refer to Figure 4(b). *The principle of stratification, well-known in statistics [36], states that any inherently ill-posed statistical system (e.g., adopted for third-stage cloud/cloud-shadow detection), will always achieve greater precision by incorporating the "stratified" or "layered" approach, provided that the input strata are as uniform as possible in respect of the characteristic of interest.* In general, input information strata are difficult or expensive to collect. Fortunately, discrete color strata (e.g., green-as-"*vegetation*") detected by the Stage 1's pre-classifiers in an input MS image are stable (data independent), informative and generated at no cost in terms of user's interactions and nearly no cost in computation time. It is worth noting that the statistical principle of stratification is equivalent to the popular divide-and-conquer (*dividi-et-impera*) problem solving approach.

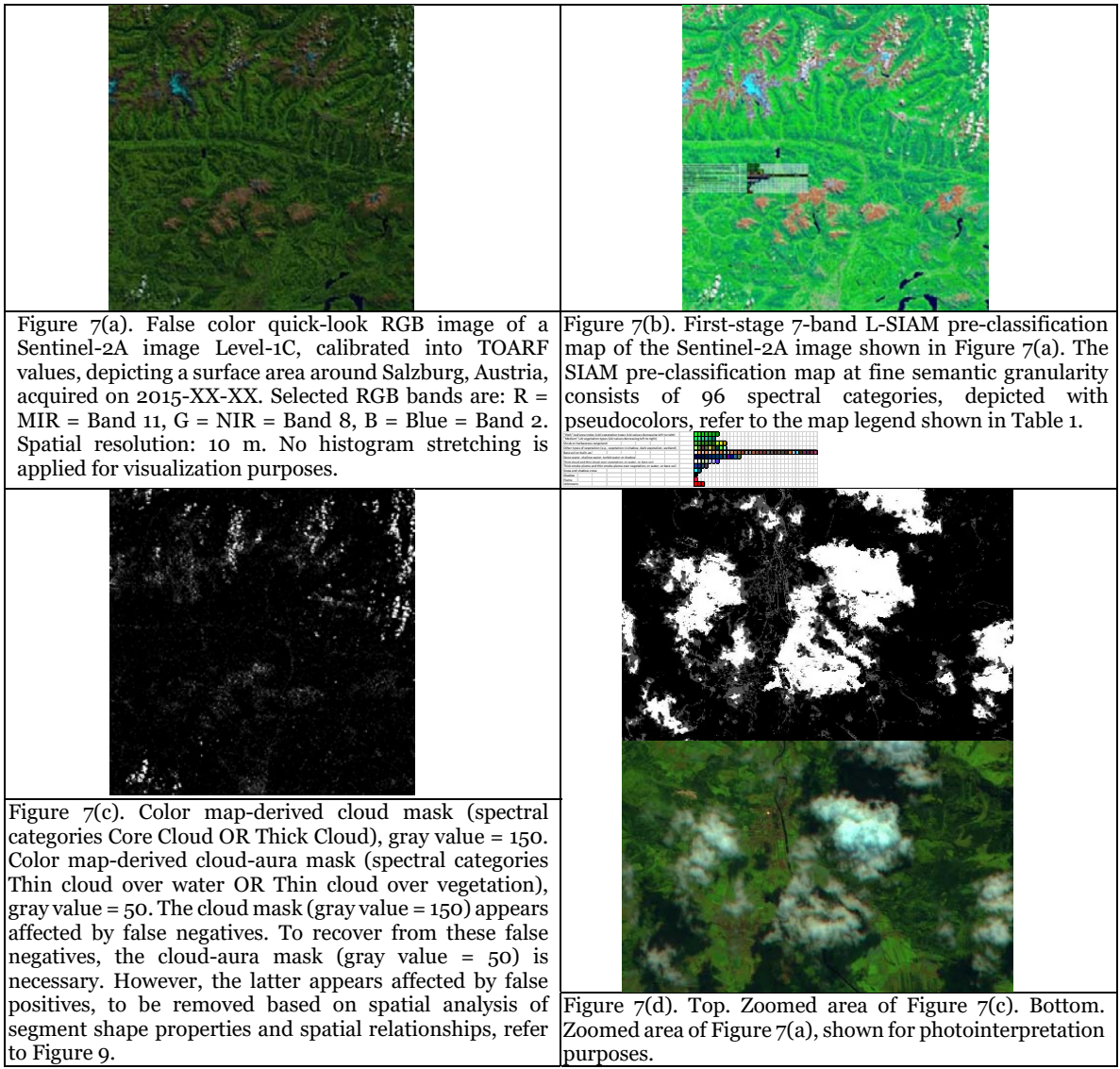

Figure 7(a). False color quick-look RGB image of a Sentinel-2A image Level-1C, calibrated into TOARF values, depicting a surface area around Salzburg, Austria, acquired on 2015-XX-XX. Selected RGB bands are: R = MIR = Band 11, G = NIR = Band 8, B = Blue = Band 2. Spatial resolution: 10 m. No histogram stretching is applied for visualization purposes.

Figure 7(b). First-stage 7-band L-SIAM pre-classification map of the Sentinel-2A image shown in Figure 7(a). The SIAM pre-classification map at fine semantic granularity consists of 96 spectral categories, depicted with pseudocolors, refer to the map legend shown in Table 1.

Figure 7(c). Color map-derived cloud mask (spectral categories Core Cloud OR Thick Cloud), gray value = 150. Color map-derived cloud-aura mask (spectral categories Thin cloud over water OR Thin cloud over vegetation), gray value = 50. The cloud mask (gray value = 150) appears affected by false negatives. To recover from these false negatives, the cloud-aura mask (gray value = 50) is necessary. However, the latter appears affected by false positives, to be removed based on spatial analysis of segment shape properties and spatial relationships, refer to Figure 9.

Figure 7(d). Top. Zoomed area of Figure 7(c). Bottom. Zoomed area of Figure 7(a), shown for photointerpretation purposes.

IV. **Stratified (driven-by-knowledge, better-posed) image segmentation, identified as Stage 2 in Figure 4(b).**

It is well known, but often forgotten in the RS common practice, that traditional (driven-without-knowledge) image segmentation is an inherently ill-posed cognitive problem [19]. On the contrary, the segmentation (partitioning) of a multi-level image is a well posed (deterministic) and automatic task to be accomplished in linear-time by a two-pass connected-component multi-level image labelling algorithm [52]. The Stage 1's output segments are image-objects (polygons) featuring low within-segment variance because their connected pixels feature the same color name, i.e., belong to the same color quantization level. Traditionally called texture elements (*texels*) or *textons* [52], these color-uniform image-objects are equivalent to the so-called *tokens* in the Marr nomenclature of the raw primal sketch [15]. In recent years, these texels have been renamed *superpixels* [53]. In series with the pre-classification Stage 1, a stratified better-posed image segmentation Stage 2 is expected to pursue so-called perceptual grouping (full primal sketch, texture segmentation) [15], i.e., it is expected to investigate the spatial organization of texels (superpixels) [37], [38].

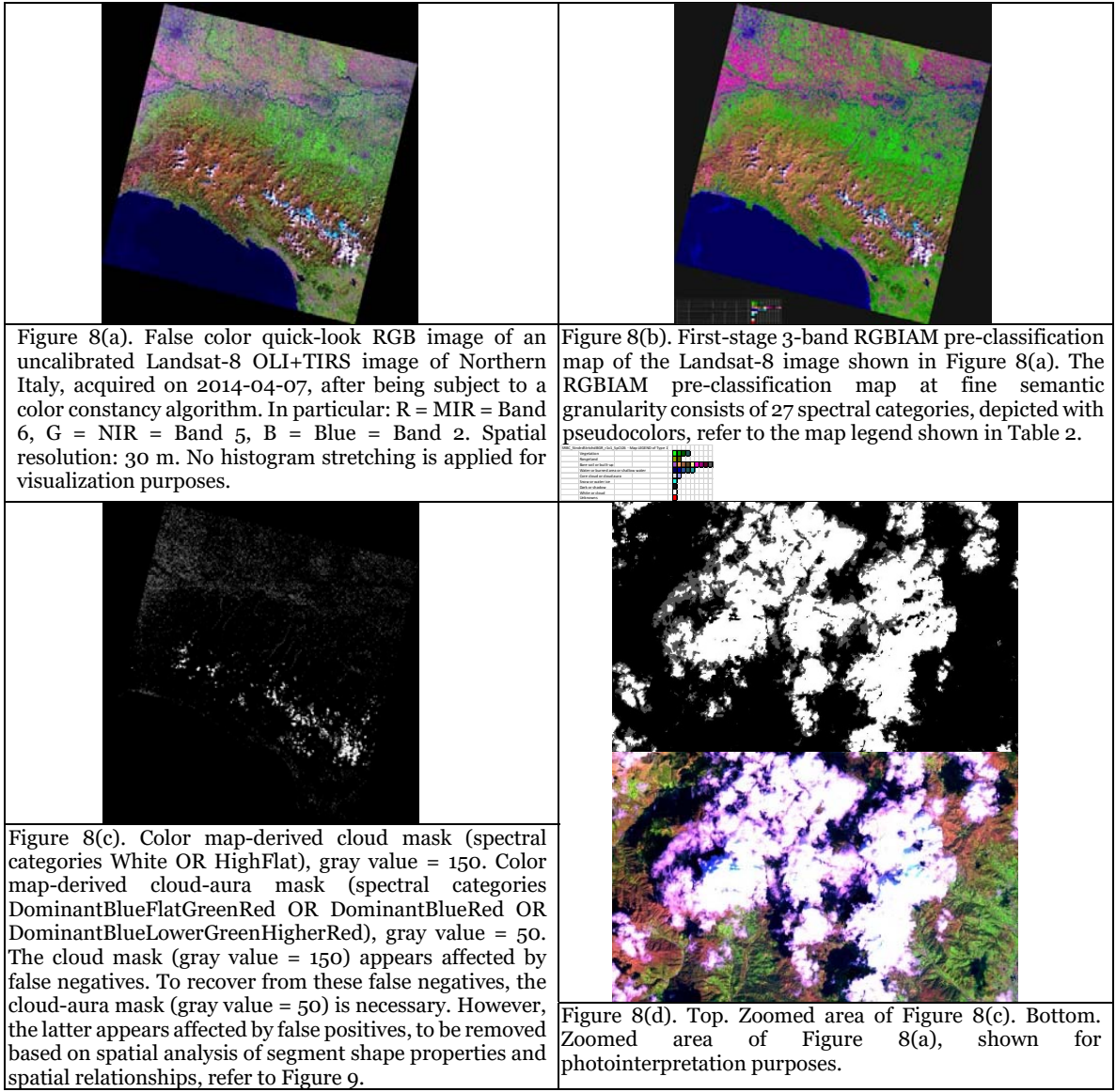

Figure 8(a). False color quick-look RGB image of an uncalibrated Landsat-8 OLI+TIRS image of Northern Italy, acquired on 2014-04-07, after being subject to a color constancy algorithm. In particular: R = MIR = Band 6, G = NIR = Band 5, B = Blue = Band 2. Spatial resolution: 30 m. No histogram stretching is applied for visualization purposes.

Figure 8(b). First-stage 3-band RGBIAM pre-classification map of the Landsat-8 image shown in Figure 8(a). The RGBIAM pre-classification map at fine semantic granularity consists of 27 spectral categories, depicted with pseudocolors, refer to the map legend shown in Table 2.

Figure 8(c). Color map-derived cloud mask (spectral categories White OR HighFlat), gray value = 150. Color map-derived cloud-aura mask (spectral categories DominantBlueFlatGreenRed OR DominantBlueRed OR DominantBlueLowerGreenHigherRed), gray value = 50. The cloud mask (gray value = 150) appears affected by false negatives. To recover from these false negatives, the cloud-aura mask (gray value = 50) is necessary. However, the latter appears affected by false positives, to be removed based on spatial analysis of segment shape properties and spatial relationships, refer to Figure 9.

Figure 8(d). Top. Zoomed area of Figure 8(c). Bottom. Zoomed area of Figure 8(a), shown for photointerpretation purposes.

## V. Stratified attentive vision third stage, identified as Stage 3 in Figure 4(b).

This is the application-specific attentive vision unit yet to be developed, where a great deal of the RTD work in Person-Month (PM) must be allocated in the project work plan (refer to further Section 1.1.8). An attentive vision third-stage battery of stratified (driven-by-knowledge), hierarchical, spatial context-sensitive sensor- and application-specific hybrid inference-based (combined physical and statistical) models for LC/LCC class-specific feature extraction and classification is identified as Stage 3, refer to Figure 4(b). Like in [31], each sensor- and class-conditional component of this second-stage battery of stratified hybrid inference-based algorithms is expected to become better/well posed in the Hadamard sense (i.e., one solution exists and is unique) [20], which also means automatic, i.e., no system's free-parameter is expected to be user-defined. It is well known that 4-D real-world objects-through-time (e.g., cars, trees, etc.) are dominated by their 4-D spatiotemporal information [27]. Hence, at the attentive vision second stage, effective exploitation of spatial information through spatial reasoning becomes mandatory. Spatial information encompasses: (a) texture (perceptual grouping) [15], (b) inter-object spatial relationships, either topological (e.g., adjacency, inclusion, etc.) or non-topological (directional, i.e., space distance or angle difference), together with (c) image-object's geometric (shape and size) features. With special emphasis on cloud detection, it is important to remember that the ATCOR-4 commercial software product partitions a scene into: (I) *clear view* (clear-sky [1], [28]), (II) hazy and (III) *cloud regions* [25]. In the words of Huang *et al.* [3], because in the troposphere (0.5 - 9 km in height) and lower stratosphere (9 – 16 km), where most clouds occur, air temperature decreases in general as altitude increases, most clouds are colder than the land or water surfaces underneath them. Such temperature differences are especially significant for mid- (6 – 9 km) to high-altitude (9 – 12 km) clouds and can be very effective in identifying such clouds. For example, because high-altitude thin cirrus clouds (see Figure 1) are not very bright spectrally, they are often very difficult to detect using non-thermal bands, but they are generally very cold, hence they can be identified relatively easily using the thermal band [3]. In addition, when the thermal band is used, like in [3], the rule-based cloud detection algorithm become much simpler than those used in cloud algorithms where no thermal band is used, e.g., [1]. In accordance with the cloud/cirrus/haze sorted sequence of mapping activities adopted in the ATCOR's decision-tree classifier (capable of generating the pixel-based output pre-classification file "_out_hcw.bsq", where acronym hcw means haze, cloud, water) [25], the proposed cloud/cloud-shadow decision-tree classifier can be structured according to the following sorted list of actions (i) to (vii).

(i) Provide the EO-IUS input variables (required) and data flags (optional), expected to be the following. (a) Spatial resolution of the input image. (b) Sun-sensor position parameters: (I) the sun zenith angle, (II) the viewing zenith angle, (III) the relative azimuth angle, see Figure 2. (c) Pre-attentive vision Stage 1: select the SIAM subversion - 7-band Landsat-like (B, G, R, NIR, MIR1, MIR2 and TIR), 4-band SPOT-like (G, R, NIR and MIR1), 4-band AVHRR-like (R, NIR, MIR1 and TIR) or 4-band VHR-like (B, G, R and NIR). (d) Attentive vision Stage 3: (I) Thermal band availability – Yes/No, (II) Cirrus band availability at 1.38 μm: Yes/No [41], e.g., band 9 in the Landsat-8 Operational Land Imager (OLI), band S4 in the future Sentinel-3 SLSTR, band M9 in NASA-NOAA Preparatory Project (NPP) - Visible Infrared Imaging Radiometer Suite (VIIRS), etc.

(ii) **Candidate core cloud detection** (see Figure 1), namely: (a) Low level clouds (0.5 – 6 km): cumulus (*Cu*), cumulonimbus (*Cb*), stratus (*St*), stratocumulus (*Sc*), nimbus (*Ni*); (b) Mid level clouds (6 – 9 km): altocumulus (*Ac*), altostratus (*As*), nimbostratus (*Ns*); (c) High level clouds (9 – 12 km): cirrus (*Ci*), Cirrocumulus (*Cc*), cirrostratus (*Cs*). Planned activities: (I) Selection of the SIAM's spectral categories, e.g., "*snow water-ice*", "*core cloud*", "*thick cloud*", as core cloud candidate image-objects, see Table 1. (II) Selection of the RGBIAM's spectral category "*white*" to detect core cloud candidate image-objects, see Table 2. (III) Select a fusion strategy to combine the SIAM's evidence with the RGBIAM's evidence to detect core cloud candidate pixels, see Figure 7 to Figure 9. (IV) Merge cloud-candidate image-objects based on spatial topological relationships: e.g., inclusion, adjacency [39], see Figure 7 to Figure 9. (V) Remove false cloud-candidate image-objects based on geometric properties, i.e., shape and size [40], see Figure 7 to Figure 9.

(iii) **Cloud anulus detection**. For example, in [3], a specific rule-based strategy is implemented in the 2-D Red – normalized TIR space to detect clouds that typically become less bright and less cold towards their edges. Planned activities: (I) Selection of the SIAM's spectral categories, e.g., "*thin cloud over water*", "*thin cloud over vegetation*", as cloud anulus candidate image-objects, refer to Table 1. (II) Merge cloud-candidate image-objects, detected in step (ii), with cloud annulus image-objects based on spatial topological relationships: e.g., inclusion, adjacency, see Figure 7 to Figure 9. (III) Remove false cloud-candidate image-objects based on geometric properties, i.e., shape and/or size, see Figure 7 to Figure 9.

(iv) **Thin cirrus detection** (see Figure 1). Thin cirrus clouds are located in the upper troposphere (6 - 9 km) or lower stratosphere (9 – 16 km) [25], hence they are typically very cold due to rapid decreases in temperature as altitude increases in the atmosphere [3]. In non-thermal wavelengths, cirrus clouds are difficult to detect, especially over land, because they are partially transparent. In [3], thin cirrus clouds over forest are identified relatively easily using the thermal band of Landsat imagery. Thin cirrus and other high altitude thin clouds over forest, which are not necessarily much brighter than the forest beneath them, but are typically very cold, are detected with a specific rule-based strategy in the 2-D Red – normalized TIR space, whereas clouds that are bright but not cold, including some near surface clouds, low fogs, and smokes, are not likely to be detectable using this cloud decision boundary. Some sensors are provided with a so-called cirrus band at 1.38 μm, e.g., Landsat-8 OLI band 9, in agreement with [41]. Water vapor dominates in the lower troposphere and usually 90% or more of the atmospheric water vapor column is located in the 0 – 5 km altitude layer. Therefore, if a narrow spectral band is selected in a spectral region of very strong water vapor absorption, e.g., around 1.38 μm or 1.88 μm, the ground reflected signal will be totally absorbed (although surface snow and water ice do appear visible in the cirrus band, based on experience, in addition to cirrus clouds), but the scattered cirrus signal will be collected by a high-altitude (> 20 km) airborne/spaceborne sensor [25]. For example, in [25], if a narrow cirrus channel around 1.38 μm exists, then two different cirrus removal strategies are adopted for water and land pixels by means of a hybrid inference system where the cirrus channel is investigated in combination with, respectively, a MIR band (water) and a Red (R) or a NIR band (land). Planned activities: (I) Investigate the SIAM's output product called Haze five-level mask. (II) If there is no thermal band or cirrus band, are there other convergence-of-evidence criteria? To be investigated.

(v) **Haze/fog detection** (see Figure 1). In the visible bands (0.35 – 0.75 μm), images contaminated by haze appear similar to those contaminated by cirrus clouds. However, in longer wavelength channels, starting from the NIR band (around 0.85 μm), haze effects are rarely visible [25]. Haze is located in the lower troposphere (0 - 3 km) as opposed to high altitude cirrus clouds, located in the upper troposphere (6 - 9 km) or lower stratosphere (9 – 16 km). For haze detection, ATCOR features two hybrid (combined rule-based plus statistical) strategies specialized to detect: (a) haze over land and, based on the visible bands Blue and Red, while the NIR band is used to exclude water pixels, and (b) haze over water, based on evidence collected from the NIR band once water pixels are masked either using spectral criteria or taking an external water map [25]. Planned activities: (I) Investigate the SIAM's output product called Haze Five-level Mask. (II) Other convergence-of-evidence criteria based on color and spatial properties?

(vi) **Smoke plume detection**. Unlike clouds, smoke plumes are not bright, but dark. Planned activities: (I) Selection of the SIAM's spectral categories, e.g., "Thin Smoke Plume over Water", "Thick Smoke Plume over Water",

"Smoke Plume over Vegetation", "Smoke Plume over Bare soil or Built-up" as smoke plume candidate image-objects, refer to Table 1. (II) Merge smoke plume-candidate image-objects with smoke plume annulus image-objects based on spatial topological relationships: e.g., inclusion, adjacency, see Figure 9. (III) Remove false smoke plume-candidate image-objects based on geometric properties, namely, shape and/or size, see Figure 9.

(vii) **_Cloud shadow detection_**, see Figure 9. In [3], the cloud height is estimated based on a statistic model-based normalized temperature and a digital elevation model (DEM) is required as ancillary input. In [1], the cloud height is assumed between 0.5 and 12 km and no DEM is employed. Planned activities: (I) Apply a physical model-based cloud shadow projection algorithm, whose input parameters are [1]: (a) the spatial location of the cloud, (b) cloud top and bottom heights, (b) the sun zenith angle, (d) the viewing zenith angle, and (e) the relative azimuth angle, see Figure 2. (II) Selection of the SIAM's spectral categories eligible as shadow candidate areas, e.g., "Water or shadow", "Vegetation in shadow", "Vegetation in water or shadow", etc., refer to Table 1. (III) Image-object matching between projected cloud shadow areas and shadow candidate image-objects can be inspired by those proposed in [3] or [26] and [49].

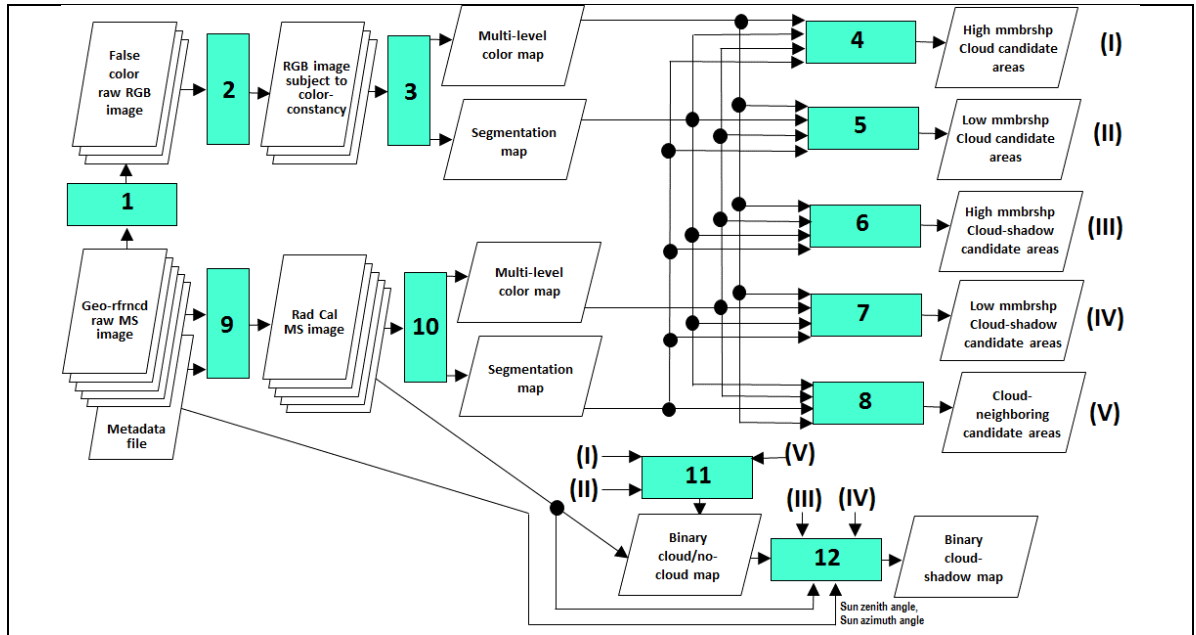

Figure 9. In agreement with the general-purpose 4-stage EO-IUS design sketched in Figure 4(b), the proposed EO-VAS architecture for automatic spatial context-sensitive cloud/cloud-shadow detection consists of a combination of deductive and inductive inference modules. (1) False color RGB channel selection, if possible (belongs to Stage 0 in Figure 4(b)). (2) Statistical self-organizing color constancy algorithm possible (belongs to Stage 0 in Figure 4(b)). (3) Prior knowledge-based RGBIAM color space discretizer (belongs to Stage 1 in Figure 4(b)). (4) AND-Candidate cloud areas (belongs to Stage 2 in Figure 4(b)). (5) NOTAND × OR-Candidate cloud areas (belongs to Stage 2 in Figure 4(b)). (6) AND-Candidate cloud-shadow areas (belongs to Stage 2 in Figure 4(b)). (7) NOTAND × OR-Candidate cloud-shadow areas (belongs to Stage 2 in Figure 4(b)). (8) Candidate cloud neighboring areas (belongs to Stage 2 in Figure 4(b)). (9) TOARF/SURF radiometric calibration (belongs to Stage 0 in Figure 4(b)). (10) Prior knowledge-based SIAM color space discretizer (belongs to Stage 1 in Figure 4(b)). (11) Spatial modeller (belongs to Stage 3 in Figure 4(b)): Clouds detected from candidate core-cloud and cloud neighboring areas. (12) Spatial modeller (belongs to Stage 3 in Figure 4(b)): Physical model-based cloud shadow detection.

To recapitulate, the original implementation of an innovative four-stage EO-IUS, sketched in Figure 4(b), suitable for cloud/cloud-shadow detection, in agreement with Figure 9, consists of several software modules that already exist. In this EO-IUS instantiation, the aforementioned low-level vision Stage 0, Stage 1 (including the RGBIAM and SIAM low-level vision expert systems, identified as processing blocks 3 and 10 in Figure 9) and at least part of Stage 2 are already available. Only the context-sensitive attentive vision Stage 2 (in part) and Stage 3 have to be implemented on an application-specific basis. The aforementioned Stage 2 (in part) and Stage 3, yet to be developed, can be identified as the two spatial modelling blocks 11 and 12 shown in Figure 9. Noteworthy, the proposed Stage 3's cloud/cloud-shadow map legend, refer to the aforementioned points (ii) to (vii), is more informative than those proposed by alternative approaches, see Table 3.

### 1.1.3 *System integration, quality assessment and comparison of alternative solutions*

Since the proposed four-stage EO-IUS design, see Figure 4(b), shall be implemented in agreement with Figure 9, where only the two processing blocks involved with spatial reasoning, identified as spatial modellers 11 and 12, are yet to be developed, then the project's RTD software activity plan focuses on the application-specific high-level vision Stage 3. In recent years, the proposed four-stage EO-IUS architecture has been implemented, integrated and tested in a variety of application-specific domains [8], [9], [12]-[14], [18], [35], [40]-[43], [50], [51]. Hence, based on past experience, the integration of a new application-specific Stage 3 with pre-existing Stages 0 to 2 is expected to be straightforward. The core of the RTD software project will focus on the development of the new high-level vision Stage 3.

Table 3. List of state-of-the-art cloud/cloud-shadow detectors in the spaceborne MS image domain.

| Paper | Sensor | Spatial resolution | MS bands | Thermal band | Rad. Cal. | Ancillary data | No-cloud/cloud-shadow LC classes Inductive (statistical) / Deductive (physical model-based) / Hybrid - LC classes | Pixel/object-based | Cloud Inductive/ Deductive / Hybrid (Deductive + Inductive) | Pixel/object-based | Cloud-shadow Pixel/object-based projection from shadow | Cloud eight estimation | Spatial search of cloud-shadow pixels | Inductive/ Deductive / Hybrid (Deductive + Inductive) |
|---|---|---|---|---|---|---|---|---|---|---|---|---|---|---|
| [1] Luo et al. | MODIS | 250 m | B1-B7, from Blue to MIR | N | TOARF or SURF | N | Deductive – LC classes: Bare soil, Vegetation, Snow/ice, Water | Pixel | Deductive | Pixel | Pixel | Assumed ranging from 0.5 km to 12 km | N | Deductive |
| [2], SPARC | AVHRR | 1.1 km | B1-B4, from Red, NIR, MIR to TIR | Y | TOARF and °K (for the thermal band) | N | Deductive – LC classes: Water, Snow | Pixel | Deductive | Pixel | Pixel | Y (based on the TIR band) | Y (in a 4-adjacency neighborhood) | N |
| [3], Huang et al. | Landsat-4/5/7 | 30 m | TM1-TM7, from Blue to TIR | Y | TOARF and °K (for the thermal band) | DEM | Deductive – LC classes: Water, Dark soil | Pixel | Hybrid (Deductive + Inductive, e.g., tile-based forest peak dtctn) | Pixel + contextual (for cloud boundary dtctn) | Pixel | Y (based on the TIR band) | Y (in the vicinity of predicted shadow pixels) | Inductive |
| [4], ACCA | Landsat-4/5/7 | 30 m | TM1-TM7, from Blue to TIR | Y | TOARF and °K (for the thermal band) | N | Deductive – LC classes: Desert soil, Snow | Pixel | Hybrid (Deductive + Inductive, e.g., image-wide histogram-based) | Pixel + contextual (for filling cloud holes) | N | N | N | N |
| [5] | SPOT-4 | 20 m | B1-B4, from Green, RED, NIR to MIR | N | Green-MIR intercalibration | N | N | N | Inductive | Pixel + object-based | Object | N | Y | Inductive |
| [25], ATCOR-4 | Airborne, spaceborne | Any | B, G, R, NIR, MIR1, MIR2 and Cirrus band (depending on the available spectral channels) | N | TOARF or SURF or Spctrl albedo | N | Deductive – LC classes: Water, Land, Haze, Snow/Ice (depending on the available spectral channels) | Pixel | Hybrid (Deductive + Inductive, e.g., image-wide histogram-based) | Pixel | N | N | N | Hybrid (Deductive + Inductive, e.g., image-wide histogram-based) |
| [26], [49] Zhu and Woodcock, Fmask | Landsat-7 | 30 m | TM1-TM7, from Blue to TIR | Y | TOARF and °K (for the thermal band) | N | Deductive – LC classes: clear land, clear water, snow. | Pixel | Deductive | Pixel + contextual (for isolated map pixel removal) | Pixel + object-based (derived by segmenting the cloud layer) | Y (based on the TIR band) | Y | Hybrid |
| EO-IUS proposed in the present proposal | Airborne, spaceborne | Any | B, G, R, NIR, MIR1, MIR2 and Cirrus band (depending on the available spectral channels) | Y | TOARF or SURF and °K (for the thrml band) | N | Hybrid – LC classes: Water, Shadow, Bare soil, Built-up, Vegetation, Snow or ice, Fire, Outliers, also refer to [30], [31]. | Pixel and object-based, refer to Section 3 | Hybrid | Pixel + object-based, refer to Section 3 | Pixel + object-based, similar to [26], [49] | Y, inspired by [3] and [26], [49] | Y, inspired by [3] and [26], [49] | Hybrid, inspired by [3] and [26], [49] |

The test/validation phase of the novel high-level classification Stage 3 is designed as follows.
 (i). A set of independent metrological Q2IOs, to be community-agreed upon, is selected from the existing literature [8], [9], refer to Section 1.1.1.
(ii). The statistically valid and spatially consistent probability sampling protocol for accuracy assessment of a fine-resolution thematic map, proposed in [42], is adopted, in contrast with non-probability sampling strategies traditionally employed in the RS common practice. In [42], both pixel-based thematic Q2Is (TQ2Is) and polygon-based Spatial Q2Is (SQ2Is), provided with a degree of uncertainty in measurement, are estimated in agreement with the GEO's QA4EO guidelines [7]. Proposed TQ2Is include the popular overall accuracy (OA), user's accuracies and producer's accuracies [3], [4]. For example, in [3], reference pixels belonging to three target LC classes, namely, cloud, cloud-shadow and clear-sky surfaces, were selected via photointerpretation. Finally, the per-pixel OA value plus per-class omission and commission errors were assessed. The same pixel-based TQ2Is estimated in [3] were also adopted in [4], [26] and [49] (refer to Table 3), without any estimation of their degree of uncertainty in measurement as a function of the reference sample size. In the present project proposal, in addition to polygon-based SQ2Is, which are omitted in [3], [4], [26] and [49], pixel-based TQ2Is must be provided with a degree of uncertainty in measurement, to comply with the GEO's QA4EO guidelines [7]. For example, typical USGS classification project requirement specifications are: OA $\in$ [0, 1] $\pm$ $\delta$ fixed equal to 0.85 $\pm$ 2% [44], where $\pm$ $\delta$ is the degree of uncertainty in measurement. A typical USGS target per-class classification accuracy, OA,c $\in$ [0, 1] $\pm$ $\delta_c$, c = 1,. . ., C, where C is the total number of target LC classes, is fixed about equal to 70% $\pm$ 5% [7]. According to Lunetta and Elvidge [45], if the desired level of significance $\alpha$ = 0.03 and C = 3, say, cloud, cloud-shadow and clear-sky surfaces, then the level of confidence (1 – $\alpha$/C) = 0.99 and $\chi2$(1, 1 – $\alpha$/C) = 6.63. In this case, if OA,c = 85%, and $\delta_c$ = $\pm$2% with c = 1, 2, 3, are the target accuracy values, then the required reference sample set size (SSS) per class c is $SSS_c$ = 2113 where c = 1, 2, 3. If OA,c = 85%, with $\delta_c$ = $\pm$5%, then $SSS_c$ = 338 with c = 1, 2, 3, and so on, refer to [42]. (III) The required reference sample must consist of multi-source images, both calibrated and non-calibrated, provided with cloud/cloud-shadow/clear-sky "truth" masks [3].

In this testing/validation scenario, two reference data sets are available free-of-cost.
(A) The Landsat-7 sensor-specific Cloud Cover Assessment Validation Data (L7CCVD) set, available for download [47], has been adopted as the reference sample in related works, like [4], [26] and [49]. Consisting of 180 Landsat-7 images, provided with radiometric calibration parameters, it covers the full range of global environments and cloud conditions. Manually selected cloud masks per reference scene and cloud-shadow masks for few reference scenes are available.
(B) Landsat-8 standard Level 1 products cirrus and non-cirrus cloud masks, encoded at the pixel level as high/medium/low confidence, can be downloaded [48]. Unfortunately, beyond these two L7CCVD and Landsat-8 reference datasets, no additional multi-source reference sample set is available yet.

To summarize, according to the selected probability sampling protocol proposed in [42]: (i) target OA$\pm$ $\delta$ and OA,c $\pm$ $\delta_c$, c = 1..., C = 3 = {cloud, cloud-shadow, clear-sky}, must be specified in advance, to compute the required $SSS_c$, c = 1, 2, 3, in agreement with [45]. (ii) Each reference sample unit features a spatial type, equal to either pixel or polygon, depending on the target LC class c = 1, 2, 3. (iii) Once randomly sampled and scrutinized by the domain experts and/or potential users, reference sample polygons, if any, must be manually edited by photointerpreters, like in [3] and [4]. To conclude, in addition to the available free-of-cost L7CCVD set and Landsat-8 imagery provided with cloud and cirrus masks, a probability sampling of multi-source reference images, featuring manually edited cloud/cloud-shadow/clear-sky "truth" objects, shall be conducted independently by potential users in the product validation phase.

Alternative to the hybrid EO-IUS proposed in this RTD project, the current state-of-the-art in cloud/cloud-shadow detection, called Fmask [26], [49] (refer to Table 3), is a purely deductive program executable, suitable for Landsat images exclusively. Fmask can be downloaded from the web page: https://code.google.com/p/fmask/. It will be adopted for direct comparison with the proposed methodology.

### 1.1.4    Differences between the proposed solution and alternative existing solutions

In the machine learning literature it is well known that *inductive data learning problems*, like image segmentation (partitioning) and its dual problem, image-contour detection [19], *are inherently ill-posed in the Hadamard sense, i.e.,* their solution does not exist or is not unique [20]. Therefore, they are very difficult to solve. *To become better posed (better conditioned) for numerical treatment, inductive data learning algorithms "require a priori knowledge in addition to data"* [21] (p. 39). By definition, prior knowledge is available *in addition to* sensory data, i.e., *a priori* knowledge is data independent, although it is typically application specific. It means that *a priori* knowledge is eligible for providing initial conditions to an inherently ill-posed adaptive learning-from-examples (statistical, inductive, bottom-up, driven-by-data, driven-without-knowledge) algorithm, such that the latter (equivalent to phenotype) is conditioned to explore a neighborhood of the former (equivalent to genotype) in a solution space.

In contrast with this common knowledge, a great majority of the existing EO-IUSs, including popular geographic object-based image analysis (GEOBIA) systems, employ a driven-without-knowledge inductive learning-from-data image segmentation first stage, which always starts its data analysis from scratch, see Figure 4(a). As a consequence, the GEOBIA first stage is affected by structural drawbacks. First, image segmentation is inherently ill-posed [19], i.e., it is semi-automatic and site-specific [22]. Second, it is sub-symbolic; as such, it falls short in addressing one key principle of vision, formulated by David Marr as follows: "vision goes symbolic almost immediately, right at the level of zero-crossing (raw primal sketch in the pre-attentive vision first stage)... without loss of information" [15] (p. 343).

Due to their lack of operativeness, existing EO-IUSs are outpaced by the ever-increasing rate of collection of EO images of enhanced quality and quantity, hereafter identified as EO "big data". For example, to date, the European Space Agency (ESA) estimates as 10% or less the percentage of EO images ever downloaded by stakeholders from its EO databases.

Supported by increasing portions of the RS literature [23], [24], the thesis that, in the RS common practice, Q2IOs of existing EO-IUSs, including GEOBIA systems, score low, can be considered part of an ongoing multi-disciplinary debate, encompassing scientific disciplines like computer vision, artificial intelligence (focused on deductive inference) and machine learning (centered on inductive inference), believed to be inadequate to provide operational solutions to their ambitious cognitive goals. *This controversy may mean that, if they are not combined, inductive and deductive inference systems show intrinsic weaknesses in operational use, irrespective of their implementation*. Whereas inductive inference systems are semi-automatic and site-specific [8], [9], [20], [21], it is well known that expert systems lack flexibility and scalability to complex problems [8], [9]. To take advantage of the unique features of each and overcome their shortcomings, statistical and physical models are increasingly combined into *hybrid inference systems* [8], [9], see Figure 4(b). For example, several existing EO-IUS implementations for cloud/cloud-shadow detection adopt a hybrid inference approach, where a prior knowledge-based decision tree is included, see Table 3 [1]-[5], [25], [26], [49]. Nevertheless, *the existing cloud/cloud-shadow detectors listed in Table 3 do not satisfy the EO-IUS requirements specified in Section 1.1.1.* First, none of these algorithms is capable of mapping an input MS image whether or not it is radiometrically calibrated. Second, all of these algorithms, but one, the popular Atmospheric / Topographic Correction for airborne/spaceborne image (ATCOR) commercial software product [25], are imaging sensor-specific, i.e., they are not scalable/transferable to different sensors. Third, with the sole exception of the pixel- and object-based methods proposed in [5], [26] and [49], the remaining cloud/cloud-shadow detectors are pixel-based, i.e., they are exclusively based on imaging spectrometry; in particular, algorithms proposed in [1], [2] and [25] employ no spatial (contextual) information whatsoever. This is in contrast with the conceptual foundation of the GEOBIA community, according to which spatial (contextual) information cannot be ignored in RS image understanding when the imaging sensor's spatial resolution is ≤ 20 m, because spatiotemporal information dominates spectral (color, context-insensitive) information in the real world, as correctly pointed out by Adams *et al.* [23], [24], [27]. This unquestionable fact is so true that, in human beings, panchromatic vision is almost as effective as chromatic vision. Last but not least, *each single cloud/cloud-shadow detector listed in Table 3 is affected by specific operational limitations at the levels of understanding of the EO-IUS design and/or implementation phase* [15]. For the sake of brevity, among the algorithms listed in Table 3, let us examine in more detail the so-called "automated" algorithm for cloud and cloud shadow detection in 30 m resolution Landsat images proposed in [3]. It consists of: (a) a physical model-based (prior knowledge-based decision-tree) detection (masking) of non-vegetation LC classes, namely, dark bare soil and water, (b) an inductive histogram analysis of a bi-modal vegetation pixel distribution collected from a moving image-window, to detect dark vegetation pixels as belonging to LC class forest, (c) a temperature normalization, where the forest temperature is subtracted from the pixel's temperature, (d) a physical rule-based detection of clouds in a 2-D red band-normalized temperature space, (e) a physical model-based estimation of the cloud height, based on temperature, (f) a physical model-based surface projection of a predicted cloud shadow from a detected cloud, (g) a contextual search of a cloud shadow in the neighborhood of a predicted cloud shadow, where dark pixels are detected in the near infra-red (NIR) or medium infra-red (MIR) bands. Operational limitations of this specific workflow are that: (A) it is not automatic, but depends on several heuristic parameters to be user-defined, (B) forest pixels are required to be detected with high confidence, to estimate a mean surface temperature expected to be higher than that of clouds, (C) there is spectral confusion between snow and cloud, and between cloud shadow and water, (D) a thermal channel is considered mandatory to accomplish the cloud shadow detection, i.e., this algorithm is thermal sensor-specific, and (E) the spatial type of information primitives is pixel and never polygon, i.e., this algorithm lacks spatial information and inter-object spatial relationships.

To summarize, in agreement with Table 3, main differences between the proposed EO-VAS in comparison with existing cloud/cloud-shadow detectors can be found: (1) at the design level of system understanding. The former employs prior knowledge-based color space partitioners, such as SIAM (for radiometrically calibrated MS images) and RGBIAM (for non-calibrated RGB images) to initialize inductive data analysis algorithms, which no longer require input parameters to be user-defined based on heuristics. (2) At the level of understanding of system implementation. Original automatic algorithms for inductive color constancy and for deductive color space quantization, such as SIAM and RGBIAM, are implemented in operating mode, which encompasses linear-time and tile-streaming implementation solutions, to be capable of processing massive images in near real-time. Consequences are that, among the competing systems revised in Table 3, the proposed EO-VAS is the only one capable of satisfying the RTD software project's requirements listed in Section 1.1.1. Last but not least, the proposed low-level vision Stage 0 and Stage 1 in Figure 4(b) accomplish an automatic mapping of color images into a mutually exclusive and totally exhaustive dictionary of color names, which is user- and application-independent. Hence, this low-level CV subsystem can be employed in a large variety of EO image-derived value-adding products and services, such as those listed at the end of Section 1.1.1, e.g., content-based EO image storage/retrieval [46].

### 1.1.5 Target user communities

Accurate detection and masking of clouds and cloud shadows is a well-known low-level vision prerequisite for clear-sky RS image mosaicking/compositing [1]-[5], suitable for further retrieval of land surface variables, either quantitative or nominal [6]. For example, cloud contamination is a relevant problem in LCC analysis, because unflagged clouds may be mapped as false LCC occurrences. In practice, accurate automatic cloud/cloud-shadow detection is a necessary not sufficient condition to transform EO image big data into operational, timely and comprehensive information products and services, in compliance with the QA4EO guidelines. For example, to date a large majority of text-based EO image querying systems employs a per-image summary statistic of cloud coverage, which carries no geospatial information about the distribution of clouds. Only few spaceborne imaging sensors, such as Landsat-8 and Sentinel-2, provide a cloud quality mask. Unfortunately, the accuracy of the Landsat-8 cloud masks has been assessed to be low [48].

To recapitulate, any existing user of EO images demands for an operational cloud/cloud-shadow detector. Therefore, potential users of the proposed computer vision VAS encompass the whole academia involved with scientific applications of EO imagery, the EO image providers, ranging from space agencies (ESA, NASA, JAXA, ISRO, CNSA, DLR, CNES, etc.) to space industry, e.g., RapidEye and DigitalGlobe, which are required to augment the accessibility of their EO big data archives by increasing the quality and quantity of image quality bands (such as the Sentinel-2A and Landsat-8 Level 2 products), the EO service industry, such as Google (Earth Engine), the EO image processing commercial software toolbox developers, such as Harris (ENVI/EDL) and Trimble (eCognition), the EO service providers, encompassing both private companies or public sector institutions, such as UN-FAO, and EO image end-users, i.e., private companies or public sector organisations where EO image-derived geo-information is integrated into their operational business practices on a regular basis.

### 1.1.6 Expected benefits of the proposed EO-VAS solution

As reported in Section 1.1.3, none of the existing cloud/cloud-shadow detectors listed in Table 3 satisfies the EO-IUS requirements specified in Section 1.1.1. On the contrary, the proposed multi-source EO-VAS implementation for cloud/cloud-shadow detection, sketched in Figure 9, is expected to satisfy the Q2IOs required in Section 1.1.1. In addition, the proposed computer vision VAS instantiation belongs to a novel hybrid feedback EO-IUS architecture, shown in Figure 4(b), suitable for a variety of low- (pre-attentional) and high-level (attentional) vision applications, listed in Section 1.1.1, including content-based image storage/retrieval in big EO image databases.

### 1.1.7 Future opportunities of the proposed EO-VAS solution

As reported in Section 1.1.1, if successful, the proposed EO-VAS solution would provide the first proof-of-concept that a novel hybrid feedback EO-IUS design and implementation strategies are capable of transforming multi-source EO image "big data" into operational, comprehensive and timely information products, in compliance with the QA4EO recommendations and with several ongoing RS international programs [16], [17]. This proof-of-concept would open a huge variety of future research, educational and market opportunities, refer to points (A) to (E) listed in Section 1.1.1.

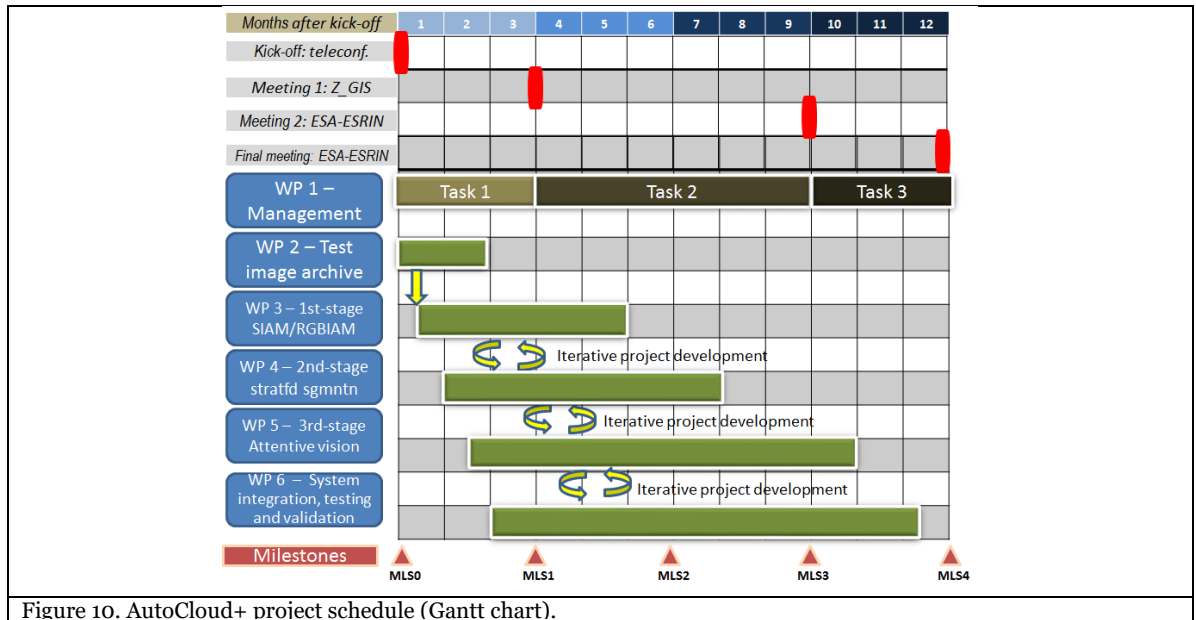

Figure 10. AutoCloud+ project schedule (Gantt chart).

### 1.1.8 Proposed approach to the work and first iteration of tasks

An iterative project development style is adopted, where the 1-year project is broken down into four 3-months iterations, corresponding to four project milestones (MLSs). In each time box a quarter of the project requirements would be addressed by completing the software life cycle for that quarter: analysis, design, code and test. At the end of each iteration predating the last iteration, the system is not expected to be put into production, but should be of production quality, to get value from the system earlier and to get better-quality feedback. In practice, the iterative project development style employing time boxing removes the critical path traditionally related to a waterfall project development style. The critical path is defined as the total time for activities on this path that is greater than that in any other path through the activity network, such that a delay in any task on the critical path leads to a delay in the project.

The project schedule (Gantt chart) is shown in Figure 10. Table 4 provides an overview of the scheduled work packages (WPs), including deliverables (Ds) and a first breakdown of manpower per WP. Intuitively, a first breakdown of costs per task can assume that costs per WP are linearly related to the person-month (PM) estimates per WP. WP1 is partitioned into Task 1 to Task 3, in agreement with the Statement of Work. WP2 (Stage 0 – Data pre-processing) to WP5 (Stage 3 – Attentive vision) correspond to Stage 0 to Stage 3 of the four-stage EO-IUS architecture sketched in Figure 4(b). Their detailed descriptions can be found in Section 1.1.2. A detailed description of WP6 (System integration, testing and validation) can be found in Section 1.1.3. An overview of the four project MLSs is provided in Table 5.

## 1.2 POTENTIAL PROBLEM AREAS

### 1.2.1 Identification of the main problem areas likely to be encountered in performing the activity

The proposed study is realistic because instances of Stage 0 to Stage 3 of the planned four-stage EO-IUS design, see Figure 4(b), were implemented, integrated and validated in recent years [8], [9], [12]-[14], [18], [35], [42], [43], [50], [51]. In Figure 9 only the two processing blocks identified as modules 11 and 12, equivalent to two spatial modellers belonging to the Stage 3 shown in Figure 4(b), are yet to be developed. Hence, based on past experience, the integration of a new application-specific Stage 3 with pre-existing Stages 0 to 2 is expected to be straightforward. The core of the RTD software project efforts will focus on the implementation of the new Stage 3's processing blocks 11 and 12 shown in Figure 9. Pre-existing system units, such as SIAM and RGBIAM, belonging to Stage 1 in Figure 4(b), are expected to require only standard maintenance. Overall, the proposed divide-and-conquer problem solving approach sketched in Figure 9, to be pursued by an iterative project development style (refer to Section 1.1.8), is capable of diluting the project technical risk. In practice, the technical risk of a major project breakdown is expected to be low or null.

Table 4. AutoCloud+ work plan, with an overview of work packages (WPs) scheduled for the proposed 12-month project, including deliverables (Ds), in agreement with Figure 10. Team member acronyms: AB = Andrea Baraldi, DT = Dirk Tiede, Sebastian d'Oleire-Oltmanns = SO.

| WP No. | WP title (with % of the WP's PM covered by each of the 3 team members; acronyms: AB = Andrea Baraldi, DT = Dirk Tiede, Sebastian d'Oleire-Oltmanns = SO). | Duration person-months (PM) – Total: 15 = 9 (AB, 75%) + 3 (DT, 25%) + 3 (SO, 25%), (hours/year @ PLUS: 1720) | Start-End, Kick-off + months, Total: 12 | Planned results, including Deliverables (Ds), in accordance with the Statement of Work, pp. 8 and 9. |
|---|---|---|---|---|
| 1 | Project Management, Quality Assessment/Validation Plan, Scientific results and software products dissemination and exploitation. | 2 | 0-12 | **Task 1** - D1.1 Service chain verification plan; D1.2 Service chain test report; D1.3 Service trial definition; **Task 2** - D1.4 Service trial report; D1.5 Service viability report; **Task 3** - D1.6 Action plan for service improvement and expansion; D1.7 Promotional material. |
|   | AB = 50%, DT = 25%, SO = 25%; Individual PMs: 1.0 + 0.5 + 0.5 = 2 | | | |
| 2 | Stage 0 (data pre-processing). Collection and radiometric calibration of multi-source EO test images and target land cover "truth", namely, cloud, cloud-shadow and clear-sky surfaces. | 1.2 | 0-2 | D2.1 Multi-source test image database provided with ground truth<br>D2.2 Multi-source test image database radiometrically calibrated into TOARF values |
|   | AB = 50%, DT = 25%, SO = 25%; Individual PMs: 0.6 + 0.3 + 0. 3 = 1.2 | | | |
| 3 | Stage 1 (pre-attentional vision - raw primal sketch). SIAM and RGBIAM installation, processing and maintenance. | 0.5 | 0.5-5 | D3.1 SIAM's products archive generated from multi-source test images |
|   | AB = 100%, DT = 0%, SO = 0%; Individual PMs: 0.5 + 0. + 0. = 0.5 | | | |
| 4 | Stage 2. Stratified image segmentation (full primal sketch). | 2 | 2-7 | D4.1 Stratified (driven-by-prior knowledge) image segmentation/texture segmentation. |
|   | AB = 70%, DT = 15%, SO = 15%; 1.4 + 0.3 + 0.3 = 2 | | | |
| 5 | Stage 3 (attentional color and spatial reasoning). Original stratified spatial context-sensitive software modules for second-stage attentive vision: development and testing.<br>(I) Stratified geometric feature extraction from image-objects.<br>(II) Estimation of inter-object spatial relationships: (i) topological (e.g., adjacency, inclusion) and (ii) non-topological (spatial distance, angle measures).<br>(III) Cloud detection (to be accomplished before cirrus and haze, refer to procedural knowledge in ATCOR [25]).<br>(IV) Cirrus detection.<br>(V) Haze detection.<br>(VI) Cloud-shadow detection. | 6.7 | 1.5-11 | D5.1 Refinement and integration of an existing software suite for stratified geometric feature extraction from image-objects.<br>D5.2 Software suite to model inter-object spatial relationships: (i) topological and (ii) non-topological.<br>D5.3 Software suite for stratified spatial context-sensitive cloud detection in multi-source test images: D5.3.1 Thermal channel available, D5.3.2 No thermal channel.<br>D5.4 Software suite for stratified spatial context-sensitive cirrus detection in multi-source test images: D5.4.1 Thermal channel available, D5.4.2 No thermal channel.<br>D5.5 Software suite for stratified spatial context-sensitive haze detection in multi-source test images: D5.5.1 Thermal channel available, D5.5.2 No thermal channel.<br>D5.6 Software suite for stratified spatial context-sensitive cloud-shadow detection in multi-source test images: D5.6.1 Thermal channel available, D5.6.2 No thermal channel. |
|   | AB = 60%, DT = 20%, SO = 20%; Individual PMs: 4.02 + 1.34 + 1.34 = 6.7 | | | |
| 6 | System integration, metrological/statistically-based quality assessment (testing) and third-party validation. | 2.5 | 3-11.5 | D6.1 Software suite for automatic cloud / cloud-shadow masking in multi-sensor multi-spectral EO images. D6.2 Test reference samples. D6.3 Validation reference samples. |
|   | AB = 60%, DT = 20%, SO = 20%; Individual PMs: 1.5 + 0.5 + 0.5 = 2.5 | | | |

### 1.2.2  *Proposed solutions to the problems identified*

An iterative project development style is adopted, where the 1-year project is broken down into four 3-months iterations. In each time box a quarter of the project requirements would be addressed by completing the software life cycle for that quarter: analysis, design, code and test. Fast prototyping of the two novel spatial modellers, identified as processing blocks 11 and 12 in Figure 9, can employ the eCognition commercial software toolbox available at the tenderer's facility.

### 1.2.3  *Proposed trade-off analyses and identification of possible limitations or non-compliances*

At the present stage of proposal, no EO-VAS methodological limitation, technical limitation or non-compliance can be reasonably foreseen.

## 1.3 TECHNICAL IMPLEMENTATION / PROGRAMME OF WORK

### 1.3.1 Proposed work logic

The work plan consists of WPs, including deliverables (Ds), summarized in Table 4. WP1 is partitioned into Task 1 (Service Verification and Service Trial Definition), Task 2 (Conduct EO service Trial) and Task 3 (Generate Action Plan & Promotional materials), in agreement with the Statement of Work. The work plan, scheduled according to the Gantt chart sketched in Figure 10, is partitioned into four quarters of the software life cycle, encompassing analysis, design, code and test. In this iterative project development style, alternative to and more flexible than a traditional waterfall project development style, there is no project's single flowchart defined beforehand where a traditional critical path can be identified, refer to Section 1.1.8.

Table 5. AutoCloud+ project overview of milestones (MLSs), in agreement with Figure 10.

| Milestone (MLS) No. | MLS title. MLS is achieved: at the end of every 1-of-4 iterations = 12 months / 4 = 3 months. | Relevant WPs involved | Expected date (months from kick-off) | Notes about WPs. |
|---|---|---|---|---|
| 0 | Kick-off | | 0 | |
| 1 | Iteration 1, ¼ of the software-life-cycle project requirements, from software analysis to design, code and test. | WP1 to WP6 | 3 | WP2 is terminated. End of Task 1. |
| 2 | Iteration 2, 2/4 of the software-life-cycle project requirements, from software analysis to design, code and test. | WP1 to WP6 | 6 | WP3 is terminated. |
| 3 | Iteration 3, 3/4 of the software-life-cycle project requirements, from software analysis to design, code and test. | WP1 to WP6 | 9 | WP4 is terminated. End of Task 2. |
| 4 | Iteration 4, 4/4 of the software-life-cycle project requirements, from software analysis to design, code and test. | WP1 to WP6 | 12 | WP5 and WP6 are terminated. End of Task 3. |

Table 6. AutoCloud+ project background: Software tools to be delivered in protected format. Acronyms: Interfaculty Department of Geoinformatics – Z_GIS of the Paris-Lodron University of Salzburg (PLUS; www.uni-salzburg.at).

| ID | Software tool to be delivered in protected format | Description | Software Language Implementation(s) | License Owner | Author(s) |
|---|---|---|---|---|---|
| 1 | RadCal | Library of sensor-specific TOARF calibrators, whose output is a MS image file in the SIAM's input file format | IDL | Andrea Baraldi | Andrea Baraldi |
| 2 | RGB cube color constancy | Automatic self-organizing RGB cube color constancy algorithm | IDL | Andrea Baraldi | Andrea Baraldi |
| 3 | SIAM | Expert system for MS calibrated image analysis (MS space partitioning into color names) and synthesis (reconstruction), equivalent to superpixel detection and quality assessment | C/C++, IDL | Andrea Baraldi | Andrea Baraldi |
| 4 | TPMLVIAS | Two-pass Multi-level Image Partition (analysis) and Piecewise Constant Continuous Image Reconstruction (synthesis, object-mean view) | C/C++ | Andrea Baraldi | Andrea Baraldi |
| 5 | RGBIAM | Expert system for RGB image analysis (RGB cube partitioning into color names) and synthesis (reconstruction), equivalent to superpixel detection and quality assessment, where the RGB image must be subject to color constancy | IDL | Andrea Baraldi | Andrea Baraldi |

### 1.3.2 Detailed procurement plan for the EO data

According to Section 1.1.3, two reference cloud/cloud-shadow image sets are available free-of-cost. First, the L7CCVD set [47], consisting of 180 Landsat-7 images provided with radiometric calibration parameters, covers the full range of global environments and cloud conditions. Manually selected cloud masks per reference scene and cloud-shadow masks for few reference scenes are available. Second, the Landsat-8 standard Level 1 products cirrus and non-cirrus cloud masks, encoded at the pixel level as high/medium/low confidence, can be downloaded [48]. In addition, radiometrically calibrated Sentinel-2A images are already available for download free-of-cost, see Figure 7. Additional test images acquired by different spaceborne sensors, e.g., RapidEye, are expected to be provided free-of-cost by potential users or are already available in the archives of this tenderer. To conclude, neither operational agreements for access to EO data nor costs for EO image purchase are envisaged.

**Automatic Spatial Context-Sensitive Cloud/Cloud-Shadow Detection in Multi-Source Multi-Spectral Earth Observation Images – AutoCloud+**

Page 18
[32] R. Benavente, M. Vanrell, and R. Baldrich, "Parametric fuzzy sets for automatic color naming," Journal of the Optical Society of America A, vol. 25, pp. 2582-2593, 2008.
[33] B. Fritzke, Some competitive learning methods, 1997. [Online]. Available: Draft document, http://www.demogng.de/. Accessed on: 28 Oct. 2014.
[34] A. Gijsenij, T. Gevers, J. van de Weijer, "Computational Color Constancy: Survey and Experiments," IEEE Trans. Image Processing, vol. X, no. X, 2010.
[35] A. Baraldi, M. Gironda, and D. Simonetti, "Operational two-stage stratified topographic correction of spaceborne multi-spectral imagery employing an automatic spectral rule-based decision-tree preliminary classifier," IEEE Trans. Geosci. Remote Sensing, vol. 48, no. 1, pp. 112-146, Jan. 2010.
[36] N. Hunt and S. Tyrrell, Stratified Sampling. Available online: http://nestor.coventry.ac.uk/~nhunt/meths/strati.html (accessed on 12 June 2014).
[37] G. M. Espindola, G. Camara, I. A. Reis, L. S. Bins, and A. M. Monteiro, "Parameter selection for region-growing image segmentation algorithms using spatial autocorrelation", International Journal of Remote Sensing, vol. 27, no. 14, pp. 3035–3040, 2006.
[38] A. Baraldi and F. Parmiggiani, ``Single linkage region growing algorithms based on the vector degree of match'', IEEE Trans. Geosci. Remote Sensing, vol. 34, no. 1, pp. 137-148, Jan. 1996.
[39] G. Christodoulou, E.G.M. Petrakis, and S Batsakis, "Qualitative Spatial Reasoning using Topological and Directional Information in OWL", In: 24th International Conference on Tools with Artificial Intelligence (ICTAI 2012). pp. 1–7. Athens (November 2012).
[40] V. B. Soares, A. Baraldi, and D. W. Jacobs, "Multi-objective software suite of two-dimensional shape descriptors for object-based image analysis," submitted for consideration for publication, IEEE J. Sel. Topics Appl. Earth Observ. Remote Sens., 2015.
[41] B. C. Gao, A. F. H. Goetz, and W. J. Wiscombe, "Cirrus cloud detection from airborne imaging spectrometer data using the 1.38 μm water vapor band," Geophysical Research Letters, vol. 20, pp. 301–304, 1993.
[42] A. Baraldi, L. Boschetti, and M. Humber, "Probability sampling protocol for thematic and spatial quality assessments of classification maps generated from spaceborne/airborne very high resolution images," IEEE Trans. Geosci. Remote Sensing, vol. 52, no. 1, Part: 2, pp. 701-760, Jan. 2014.
[43] D. Tiede, F. Lüthje, and A. Baraldi. Automatic post-classification land cover change detection in Landsat images: Analysis of changes in agricultural areas during the Syrian crisis. In E. Seyfert, E. Gülch, C. Heipke, J. Schiewe, & M. Sester (Eds.), Publikationen der Deutschen Gesellschaft für Photogrammetrie, Fernerkundung und Geoinformation (DGPF) e.V. Band 23, Potsdam. 2014.
[44] T. Lillesand and R. Kiefer, Remote Sensing and Image Interpretation. New York, NY, USA: Wiley, 1979.
[45] R. S. Lunetta and C. D. Elvidge, Remote sensing and Change Detection: Environmental Monitoring Methods and Applications. Chelsea, MI, USA: Ann Arbor Press, 1998.
[46] S. Natali and A. Baraldi, "Semantic-geospatial query of remotely sensed image archives," ESA-EUSC 2006: Image Information Mining For Security and Intelligence, EUSC Torrejon Air Base, Madrid (Spain), Nov. 27-29, 2006.
[47] USGS, Cloud Cover Assessment Validation Data. Available online: http://landsat.usgs.gov/ccavds.php#Austral (accessed on 2 Dec. 2014).
[48] V. Kovalskyy and D. P. Roy, "A one year Landsat 8 conterminous United States study of cirrus and non-cirrus clouds", Remote Sens., vol. 7, pp. 564-578, 2015.
[49] Z. Zhu C. E. and Woodcock, "Improvement and Expansion of the Fmask Algorithm: Cloud, Cloud Shadow, and Snow Detection for Landsats 4-7, 8, and Sentinel 2 Images," Remote Sensing of Environment, in press, 2015 (paper for Fmask version 3.2.).
[50] A. Baraldi, D. Tiede, and S. Lang, "Automatic Linear-Time Prior Knowledge-Based Multi-Level Color Analysis and Synthesis of RGB Imagery for Superpixel Detection and Quality Assessment," submitted for consideration for publication, IEEE Trans. Pattern Anal. Machine Intell., TPAMI-2015-06-0436, 2015.
[51] A. Baraldi, D. Tiede, M. Belgiu, and M. Sudmanns, Project winner of the T-Systems Big Data Challenge of the Copernicus Masters 2015: "Satellite Image Automatic Mapper™ (SIAM™)-Through-Time (SIAMT2) for spaceborne/airborne multi-spectral image time-sequence classification in operating mode and content-based image database retrieval" (Project ID 150688)". Awards Ceremony on Oct. 20, 2015 at the Satellite Masters Conference, 20-22 Oct. 2015, German Federal Ministry of Transport and Digital Infrastructure, Invalidenstraße 444, 10115 Berlin, Germany.
[52] M. Sonka, V. Hlavac, and R. Boyle, Image Processing, Analysis and Machine Vision. London, U.K.: Chapman & Hall, 1994.
[53] R. Achanta, A. Shaji, K. Smith, A. Lucchi, P. Fua, and S. Susstrunk, "SLIC superpixels compared to state-of-the-art superpixel methods," IEEE Trans. Pattern Anal. Machine Intell., vol.. 6, no. 1, pp. 1-8, vol. 6, no. 1, 2011.
[54] J. Clevers, O. Vonder, R. Jongschaap, J. Desprats, C. King, L. Prevot, and N. Bruguier, "Using SPOT data for calibrating a wheat growth model under mediterranean conditions," Agronomie, EDP Sciences, vol. 22, no. 6, pp.687-694, 2002.